# Bayesian Discovery of Linear Acyclic Causal Models

**Patrik O. Hoyer**
Helsinki Institute for Information Technology
& Department of Computer Science
University of Helsinki
Finland

**Antti Hyttinen**
Helsinki Institute for Information Technology
& Department of Computer Science
University of Helsinki
Finland

## Abstract

Methods for automated discovery of causal relationships from non-interventional data have received much attention recently. A widely used and well understood model family is given by linear acyclic causal models (recursive structural equation models). For Gaussian data both constraint-based methods (Spirtes et al., 1993; Pearl, 2000) (which output a single equivalence class) and Bayesian score-based methods (Geiger and Heckerman, 1994) (which assign relative scores to the equivalence classes) are available. On the contrary, all current methods able to utilize non-Gaussianity in the data (Shimizu et al., 2006; Hoyer et al., 2008) always return only a single graph or a single equivalence class, and so are fundamentally unable to express the degree of certainty attached to that output. In this paper we develop a Bayesian score-based approach able to take advantage of non-Gaussianity when estimating linear acyclic causal models, and we empirically demonstrate that, at least on very modest size networks, its accuracy is as good as or better than existing methods. We provide a complete code package (in R) which implements all algorithms and performs all of the analysis provided in the paper, and hope that this will further the application of these methods to solving causal inference problems.

## 1 INTRODUCTION

Causal relationships have a fundamental status in science because they make it possible to predict the results of external actions and changes (i.e. interventions) applied to a system. Although the preferred way of inferring causality is through randomized controlled experiments, it is often the case that such experiments cannot be performed. In such situations, methods for inferring causal relationships from passively observed data would be very useful.

For continuous-valued data, one well-known approach to this so-called 'causal discovery' problem is to fit a *linear, acyclic* causal model (recursive structural equation model, SEM) to the observed data. Although it is seldom possible to completely rule out the existence of hidden confounding variables, a first analysis is typically performed using methods which assume causal sufficiency. In this vein, we will here restrict ourselves to linear, acyclic models without hidden variables and without hidden selection effects.

For Gaussian models, it is well-known that there exist Markov equivalence classes of directed acyclic graphs (DAGs) over the observed variables, such that two DAGs in the same equivalence class represent the exact same set of observed distributions. Thus constraint-based methods such as PC and IC (Spirtes et al., 1993; Pearl, 2000) output an inferred equivalence class rather than a DAG. Score-based methods (Geiger and Heckerman, 1994) return scores (such as marginal likelihoods or posterior probabilities) over the equivalence classes, and are thus able to better express uncertainty as regards to what the true equivalence class is.

Recently, however, it has been shown that if the observed data is *non-Gaussian*, causal discovery methods can go beyond the Markov equivalence classes of the Gaussian case, and in the large sample limit infer the full DAG (Shimizu et al., 2006). It has also been shown how to combine the Gaussian and the non-Gaussian approaches to yield as much information as possible in the mixed case (Hoyer et al., 2008). These methods, however, always return only a single DAG or a single equivalence class, and thus cannot convey the degree to which the answer may be uncertain. In other words, score-based methods for the non-Gaussian case have been lacking.

In this paper, we introduce a Bayesian score-based method for causal discovery of linear, acyclic causal models under the assumption of causal sufficiency. Our approach is insensitive to the particular form of the distributions: When the observed data is close to Gaussian the method approximates the Gaussian score-based approach of Geiger and Heckerman (1994) whereas for increasingly non-Gaussian data the method is able to differentiate between the DAGs within the Markov equivalence classes. The method thus provides a unified and relatively robust method for linear causal discovery.

The paper is structured as follows: In Section 2 we formally define the model family under consideration, followed by a derivation of the marginal likelihood of the observed data in Section 3. Then, in Section 4, we discuss the important topic of choosing a model family (and priors) for the densities of the disturbances, after which we present empirical results in Section 5 based on simulations with both artificial and real data. Some discussion on how to tackle the computational problems involved in learning larger networks is provided in Section 6 while conclusions are given in Section 7.

## 2 MODEL

The data-generating models we are considering can be described formally as follows:

1. There is a one-to-one mapping between the observed variables $x_i$, $i = \{1 \ldots n\}$ and the nodes of a directed acyclic graph (DAG) $\mathcal{G}$.

2. The value assigned to each variable $x_i$ is a *linear function* of the values already assigned to the variables constituting its parents in $\mathcal{G}$, plus a 'disturbance' (noise) term $e_i$, plus a constant term $c_i$, that is
$$x_i := \sum_{j \in \text{pa}(i)} b_{ij} x_j + e_i + c_i, \quad (1)$$
where $j \in \text{pa}(i)$ if and only if there is a directed edge in $\mathcal{G}$ from the node corresponding to $x_j$ to that corresponding to $x_i$. (Thus, variables corresponding to nodes with no parents are drawn first, their children next, and so on until all variables have been assigned values.)

3. The disturbances $e_i$ are all continuous random variables with arbitrary densities $p_i(e_i)$, and the $e_i$ are mutually independent, i.e. $p(e_1, \ldots, e_n) = \prod_i p_i(e_i)$.

The observed data consists of $N$ observed data vectors $\mathbf{x}^1, \ldots, \mathbf{x}^N$ (each containing all the variables $x_i$), each data vector having been generated by the above process, with the same graph $\mathcal{G}$, the same coefficients $b_{ij}$, the same constants $c_i$, and the disturbances $e_i$ sampled independently from the same densities $p_i(e_i)$.

Note that the above assumptions imply that the observed variables are causally sufficient (Spirtes et al., 1993), i.e. there are no unobserved confounders (Pearl, 2000).

## 3 INFERENCE

### 3.1 Basic Setup

Given the observed dataset $\mathcal{D} = \{\mathbf{x}^1, \ldots, \mathbf{x}^N\}$ our task is to infer the data-generating model; in particular we are interested in the network structure (the DAG) that generated the data.

Here we take a very straightforward Bayesian approach to inference of the DAG. Denoting the different possible DAGs by $\mathcal{G}_k$, $k = \{1, \ldots, N_g\}$ (where $N_g$ denotes the number of different DAGs on $n$ variables) and grouping all the parameters (i.e. the coefficients $b_{ij}$, the constants $c_i$, and the disturbance densities $p_i(e_i)$) into $\theta$, we will base our method on the marginal likelihood of the data conditional on the different graphs:

$$p(\mathcal{D} \mid \mathcal{G}_k) = \int p(\mathcal{D} \mid \theta, \mathcal{G}_k) \, p(\theta \mid \mathcal{G}_k) \, d\theta. \quad (2)$$

When it is reasonable to assume that the true data-generating model is in this model family the marginal likelihoods can be converted to posterior probabilites through Bayes rule

$$P(\mathcal{G}_k \mid \mathcal{D}) = \frac{p(\mathcal{D} \mid \mathcal{G}_k) \, P(\mathcal{G}_k)}{p(\mathcal{D})}, \quad (3)$$

where $p(\mathcal{D})$ is a constant (with respect to $k$) which simply normalizes the distribution. Here $P(\mathcal{G}_k)$ is the prior probability distribution over DAGs and incorporates any domain knowledge we may have. In this paper we for simplicity assume a uniform prior over all DAGs; there is no need to explicitly penalize for model complexity because such penalization is implicit in computing the marginal likelihood, see Section 3.3.

### 3.2 Likelihood and Priors

To compute an approximation to (2) we need to specify $p(\mathcal{D} \mid \theta, \mathcal{G}_k)$ and $p(\theta \mid \mathcal{G}_k)$, and then approximate the integral somehow. Here, $p(\mathcal{D} \mid \theta, \mathcal{G}_k)$, the probability density of the data given the full model (structure and parameters), is fully specified by the definition of the model in Section 2:

$$p(\mathbf{x} \mid b_{ij}, c_i, p_i, \mathcal{G}_k) = \quad (4)$$

$$\prod_i p_i(x_i - \sum_{j \in \mathrm{pa}(i)} b_{ij} x_j - c_i),$$

because the $x_i$ are just an affine transformation of the $e_i$ and furthermore the transformation is one of *shearing* so the determinant is equal to one. (Note that because of acyclicity the matrix $\mathbf{B}$ containing the coefficients $b_{ij}$ is permutable to lower-triangular for a causal ordering of the variables.) This, together with the assumption of i.i.d. data, $p(\{\mathbf{x}^1, \ldots, \mathbf{x}^N\}) = \prod_m p(\mathbf{x}^m)$, fully specifies the probability density of the data given the model structure and parameters.

The prior over the parameters, $p(\theta \mid \mathcal{G}_k)$, remains for us to specify. Optimally, it would reflect our actual beliefs concerning typical interaction strengths and typical distributions, in whatever specific problem setting the method is applied on. Of course, such beliefs can be quite vague and difficult to quantify precisely. As is common in Bayesian approaches we thus select the priors based on considerations of conceptual and computational simplicity, while trying to make sure that they reflect what could be considered reasonable expectations in real-world applications. Because we want our method to be indifferent to the location and scale of all variables, we standardize them to zero mean and unit variance before applying our method. We are thus left with determining reasonable priors in such cases.

At first sight, it would thus seem reasonable to design $p(\theta \mid \mathcal{G}_k)$ to explicitly enforce the zero-mean, unit-variance property of the observed variables. This, however, would create complicated dependencies between the components of $\theta$. Therefore we instead select relatively simple independent priors which, while not enforcing standardized variables, are not incompatible with this case: We use a simple zero-mean, unit-variance Gaussian prior for the coefficients $b_{ij}$ while, because of the standardization of the variables in the inference procedure, the estimated coefficients $c_i$ can all be set to zero and thus neglected. We discuss the parametrization of (and the related prior on) the densities $p_i(e_i)$ in Section 4.

### 3.3 Approximating the Integral

Finally, we need a way to approximate the multidimensional integral (2) over the space of parameter values $\theta$. A number of possibilities exist, including analytical and sampling approaches (Gelman et al., 2004). Our main implementation simply uses the Laplace approximation: We use nonlinear optimization techniques to find the maximum of the integrand, compute the curvature (the Hessian matrix) of the logarithm at this point, and approximate the value of the integral by that of a scaled Gaussian with the same maximum value and the same log-curvature. This is a good approximation if the posterior distribution is unimodal and if the number of samples is enough for the posterior to be well approximated by a Gaussian.

To validate the Laplace approximation, and as an alternative method of computing the marginal likelihood, we have also implemented Markov Chain Monte Carlo (MCMC) sampling using the Metropolis algorithm (Metropolis et al., 1953) for the 'GL' density parametrization (see Section 4). Results were qualitatively similar and are not shown; the method can easily be tested using our freely available software package.

Note that model selection by marginal likelihood performs a kind of automatic Occam's razor: Unnecessarily complex graphs $\mathcal{G}_k$ (with more edges than needed to explain the data) are penalized as compared to simple models; see e.g. (Rasmussen and Ghahramani, 2001). Although not strictly required for linear-non-Gaussian inference (Shimizu et al., 2006), such model selection is essential in the linear-Gaussian setting (Spirtes et al., 1993).

### 3.4 Parameter Modularity

Two important practical obstacles remain: the high dimensionality of the space of the parameters $\theta$ and the super-exponential number of DAGs with respect to the number of observed variables. Even with a relatively small number of parameters for each disturbance density $p_i(e_i)$ (see Section 4) the size of the parameter vector can quickly become an obstacle to efficient inference. Fortunately, we can naturally take advantage of parameter independence and parameter modularity (Geiger and Heckerman, 1994; Heckerman et al., 1995). Dividing up the parameters into 'components' or 'families' corresponding to the observed variables, each component containing the disturbance distribution at that node and the coefficients on all edges pointing into the node, the prior can be decomposed into independent factors, one for each component. This decomposition will then also hold in the posterior over the parameters. Thus the integral (2) can be evaluated in terms of smaller independent parts.

The above factorization also goes some way to alleviating the explosive growth in the number of DAGs, because the DAGs simply combine a much smaller number of components in different ways. Thus, once the scores for all possible $n\,2^{(n-1)}$ components have been computed (each of the $n$ nodes can have any subset of the other variables as its parents), the score for any DAG can easily be obtained by combining the relevant components. This makes it feasible to score all DAGs for up to about 6 observed variables. This issue is explored further in Section 6.

## 4 MODELING THE DENSITIES

Perhaps the most important modeling decision concerns how to parametrize the densities $p_i(e_i)$, and what priors to use. Since we want our model to be able to represent both Gaussian and non-Gaussian data, the density family should include the Gaussian as a special case. It should be relatively easy to compute the density at any point given the parameters, and it should also be relatively easy to find reasonable starting values for the parameters in the required optimization. We have implemented two quite basic parametrizations: a simple two-parameter exponential family distribution combining the Gaussian and the Laplace distributions, and a finite Mixture of Gaussians density family.

The density of the first parametrization (which we will term 'GL' for short) is given by

$$p_i(e_i) = \exp\left(-\alpha_i|e_i| - \beta_i e_i^2\right)/Z_i, \quad (5)$$

where we require $\beta_i > 0$. The normalization constant can be evaluated as

$$Z_i = 2\int_0^\infty \exp\left(-\alpha_i e_i - \beta_i e_i^2\right) de_i \quad (6)$$

$$= \frac{\sqrt{\pi}}{\sqrt{\beta_i}} \exp\left(\frac{\alpha_i^2}{4\beta_i}\right)\left(1 - \operatorname{erf}\left(\frac{\alpha_i}{2\sqrt{\beta_i}}\right)\right)$$

Note that this distribution is always symmetric with respect to the origin. (It would be relatively simple to extend the family to also include skewed distributions, by adding a parameter shifting the absolute value term with respect to the quadratic term, but for simplicity we have not yet implemented this. Note that skewed distributions can easily be represented by the Gaussian Mixture model given below.) The parameters $\alpha_i$ and $\beta_i$ constitute the parameters in $\theta$ representing the density of the disturbance at node $i$. We use a Gaussian prior for $\alpha_i$ and $\log \beta_i$, giving a mix of super-Gaussian, sub-Gaussian, and close to Gaussian distributions. Random draws from this distribution are shown in Figure 1a.

The second density family we have used is the finite component Mixture of Gaussians ('MoG') density given by

$$p_i(e_i) = \sum_j \pi_j \mathcal{N}(\mu_j, \sigma_j^2), \quad (7)$$

where for each $j$ we have $\pi_j \geq 0$ and additionally we require $\sum_j \pi_j = 1$. It remains to put a prior density on the parameters $\pi_j, \mu_j$, and $\sigma_j^2$. For computational simplicity in the optimization procedure, we choose to operate with a parametrization using unrestricted $\gamma_j$ and represent $\pi_j$ through

$$\pi_j = \frac{\exp(\gamma_j)}{\sum_i \exp(\gamma_i)}, \quad (8)$$

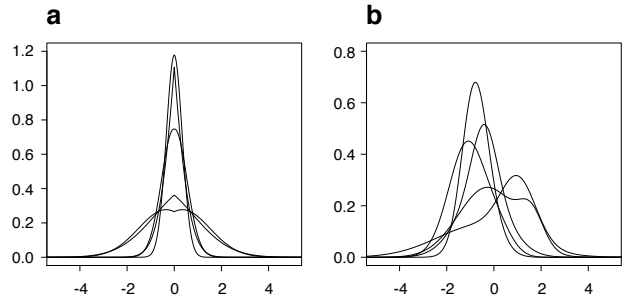

Figure 1: Families of densities for modeling the disturbance distributions. **(a)** Five random draws from the prior using the GL density parametrization given by (5). **(b)** Five random draws from the prior using the MoG parametrization (7). Note that in both cases the parametrization can represent a variety of both Gaussian and non-Gaussian distributions.

and the $\gamma_j$ are mutually independent and normal. Likewise, the priors over $\mu_j$ and $\log \sigma_j$ are Gaussian. This prior works reasonably well if the residuals are normalized to zero mean and unit variance before fitting the densities. In most of the empirical work we have simply used a two-component mixture. Random draws from our prior distribution are shown in Figure 1b.

## 5 EMPIRICAL STUDY

Please note that the full code to run all algorithms and produce all of the figures in this paper is available at:

http://www.cs.helsinki.fi/group/neuroinf/
lingam/bayeslingam/

### 5.1 Evaluation Measures

We seek to compare the accuracy of our method, termed 'BayesLiNGAM', to the PC algorithm (Spirtes et al., 1993), the algorithm of Geiger and Heckerman (1994) (termed 'GH' in what follows), and to the LiNGAM method (Shimizu et al., 2006). The comparison is a bit complicated due to the fact that they all output somewhat different structures: Whereas PC basically returns a single Markov equivalence class[1] and LiNGAM returns a single DAG, the method of Geiger and Heckerman essentially returns posterior probabilities over all equivalence classes and BayesLiNGAM returns such probabilities over all DAGs. Because we know the true generating graph in all of our experiments (i.e. these are 'gold standard' experiments), the measures we use to evaluate success will be 'loss functions' that quantify the degree of discrepancy between the predictions of the various methods and the true graph.

---

[1] Actually, due to unavoidable statistical errors, in some cases the output of PC does not correspond to any equivalence class.

The first and simplest measure is the **binary loss** which assigns a unit loss whenever the true graph was not the 'best guess' of the method in question. For LiNGAM this best guess is naturally the output DAG, whereas for `BayesLiNGAM` the best guess is the DAG with highest probability. Since PC and GH do not differentiate within the equivalence classes, we define their best guess DAG to be a random DAG from the output equivalence class and equivalence class with maximum probability, respectively. Similarly, as a second measure we define a **class loss** which quantifies how often the wrong equivalence class was guessed. In this case, the output of LiNGAM and `BayesLiNGAM` must be mapped to the respective equivalence class before testing for equality with the equivalence class of the true graph.

The above measures indicate how good or bad our best guess is, but do not encourage the methods to signal uncertainty. Since `BayesLiNGAM` and GH output probabilities rather than single graphs we would like to evaluate how well these probabilistic forecasts work. This setting is quite standard, and several measures (technically known as 'proper scores') exist that reward both forecasting accuracy and honesty; for a review see e.g. (Dawid, 1986). Here we use the well-known **logarithmic loss** given by the negative logarithm of the probability assigned to the true graph, as well as the **quadratic loss** defined as the sum of squares of the difference between the assigned probability vector and the vector $(0, \ldots, 0, 1, 0, \ldots, 0)$ where the single 1 is assigned to the true graph. For these loss functions, the GH method is assumed to divide up the probability for a Markov equivalence class equally among its constituent DAGs, while the PC method is assumed to divide up all probability mass between the DAGs representing the selected equivalence class. LiNGAM is simply taken to set all probability mass on the selected DAG.

### 5.2 Synthetic Data Simulations

To obtain statistically significant results on the accuracies of the different methods it is crucial to have a quite large set of test cases, which is most easily achieved in simulations with completely synthetic data. In this subsection we perform precisely such experiments, where we control all aspects of the data-generating process. Then, in the next subsection, we explore the use of real data in evaluating the relative strenghts of the algorithms.

Because of the relatively heavy computational burden of the `BayesLiNGAM` method, the need to test the method on everything from small samples to relatively large samples, and the need to have a large number of test cases for each setting of the test parameters, we start with a simple simulation of the most elementary of settings: the case of just two observed variables. Thus there are just three possible DAGs ($x_1 \quad x_2$, $x_1 \rightarrow x_2$, and $x_1 \leftarrow x_2$) of which the last two are Markov equivalent to each other. For this case, approaches based on covariance information alone can only distinguish between the empty graph and the connected equivalence class (essentially just performing a test of independence), whereas methods which use higher-order statistics are able to distinguish all three DAGs given enough data if the variables are non-Gaussian.

We generated data by repeatedly (1) randomly selecting one of the 3 different DAGs over the observed variables, (2) drawing the coefficient corresponding to the edge (if present) from a uniform distribution on $[-3, 3]$, (3) drawing the disturbances $e_i$ from normal distributions and transforming them using a nonlinearity $f(e_i) = \text{sign}(e_i) |e_i|^q$, and (4) generating the values for the observed variables $x_1$ and $x_2$ given this model, and subsequently shifting and scaling them such that their means and variances contained no information about the underlying causal model.

The main focus of these simulations is on the behaviour of the various algorithms as a function of two parameters: the non-Gaussianity of the variables and the number of sample vectors available. Hence, the exponent $q$ was systematically varied in logarithmic steps between $\exp(-1)$ and $\exp(1)$, and the number of generated samples $N$ varied from 10 to 10,000. For each combination of $q$ and $N$, we generated 1,000 datasets in the above fashion, and ran each of the competing methods on each dataset. The results are displayed in Figure 2. The five columns give the results of our `BayesLiNGAM` method using the GL and MoG density parametrizations, the Bayesian Gaussian approach of Geiger and Heckerman (1994) as implemented in the R package 'deal' (Bøttcher and Dethlefsen, 2003), the LiNGAM algorithm of Shimizu et al. (2006), and the standard PC algorithm (Spirtes et al., 1993). The rows of the figure correspond to the different evaluation measures (see Section 5.1). For each pair of (method, measure), the figure shows the performance with respect to sample size and non-Gaussianity. Lighter shades are better (lower loss), and for each measure separately, the shades among the various algorithms are comparable.

All methods produce some errors since small correlations and non-Gaussianities are unnoticeable with limited sample sizes. Compared to the LiNGAM-algorithm, our `BayesLiNGAM` procedure works significantly better at low sample sizes, in particular for strong non-Gaussianity. This is the case both in terms of the 'best guesses' provided by the methods (as evident from the binary and class losses) and also in terms of the uncertainty expressed (as can be seen from the logarithmic and quadratic loss functions). On the other hand, compared to the Gaussian procedures (GH and PC) `BayesLiNGAM` is able to exploit non-Gaussianity while still yielding essentially the same performance in the Gaussian cases ($q = 1$, middle row of each plot). Thus it seems that our algorithm is able to combine the strong points of both Gaussian and non-Gaussian inference methods.[2]

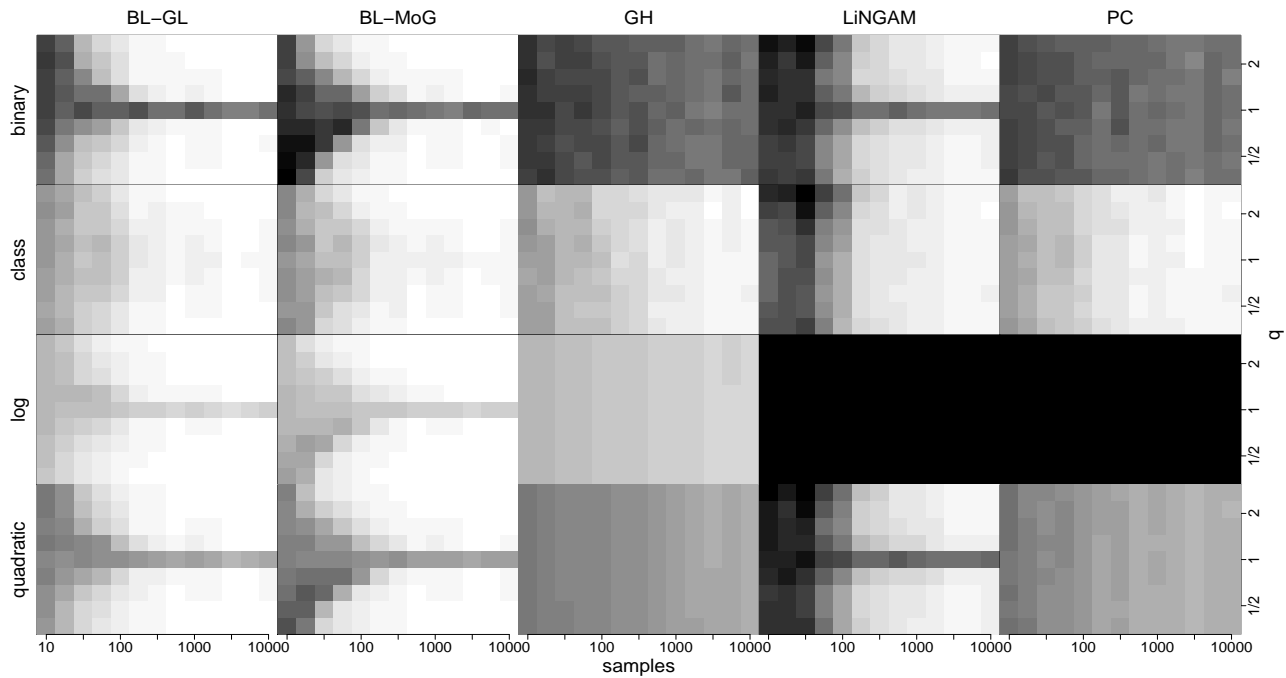

Figure 2: Results on synthetic data with 2-node DAGs. The figure shows the average values of the loss functions (rows) of each method (columns) for all combinations of sample size (horizontal axis) and non-Gaussianity (vertical axis, $q = 1$ denotes Gaussian data). The shade of gray of each square represents the average of 1000 simulations. Lighter shades indicate better results (lower loss) and shades on each row are comparable with each other (for the binary loss, black denotes 67% errors, white 0%; for the class loss black is 50% errors and white is 0%). Note that both LiNGAM and PC sometimes effectively assign zero probability to the true graph, resulting in infinite log-loss (explaining the all-black color of the respective images). See main text for details and analysis.

An interesting question that we can answer in these simulations is the degree to which `BayesLiNGAM` and the method of Geiger and Heckerman (1994) output approximately calibrated probabilities (Dawid, 1986). That is, are the relative frequencies roughly matched to the predicted probabilities? In Figure 3 we show plots indicating that this is indeed approximately the case. Of course, since both methods are based on Bayesian inference, the degree of calibration depends to a very large extent on the degree of correspondence between the priors and distribution of the test parameters, so these plots must be considered specific to the testing setting. Nevertheless, given that we did not particularly try to match the tests to the priors in the algorithms, it is reassuring that the methods still are close to calibrated.

Finally, to verify the competitiveness of our method we performed further simulations with synthetic data for slightly larger networks ($n = 3, 4$, and 5), necessarily with a less comprehensive exploration of the test parameters. For lack of space, these results are not shown, but can be reproduced using our online code. See also the discussion and example in Section 6 on extending the method to handle larger networks.

### 5.3 Simulations Based on Real Data

Although `BayesLiNGAM` did quite well in the synthetic simulations, the performance of the method in a broader context depends to some extent on the degree to which the priors in the method match the parameters of actual data-generating processes. That is, the interesting question is of course how well the different methods do on problems involving *real data*. To get a definitive answer to this, we would need a large, diverse set of representative real datasets for which the true graph is known with certainty. Unfortunately, we do not have enough such datasets to do a statistical analysis. Therefore we will use a proxy: we utilize real datasets for constructing 'realistic' networks (graphs and edge coefficients) and use true observed distributions to generate the data. (This approach could be called 're-simulating' the data.)

---

[2] Note that in this two-variable setting, LiNGAM is not too bad for Gaussian data ($q = 1$), generally getting the equivalence class correct with enough samples. This, however, does not hold in general for Gaussian data over a larger number of variables.

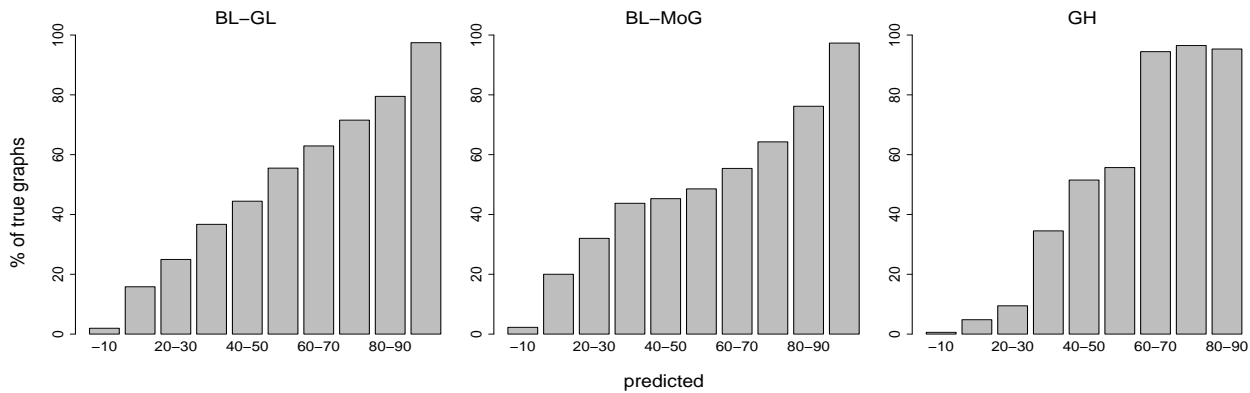

Figure 3: Calibration curves or 'reliability diagrams' (Dawid, 1986) of the three methods which give probabilistic predictions. Each bar shows the relative frequency (vertical axis) of the true graph in the set of cases where a predicted probability (horizontal axis) was given. If the methods are well calibrated these numbers should match, at least approximately.

Briefly, given a real dataset of $N$ samples over $n$ variables, we first learn a DAG using PC (randomly selecting a DAG from the resulting equivalence class). Next we estimate the edge coefficients using standard ordinary least squares regression. We then calculate the corresponding residuals, and shuffle each residual so that the residuals are mutually independent. Finally, we create 'new' data from the learned DAG, using the estimated regression coefficients and the shuffled residuals, and subsample if necessary to obtain a number of datasets of size 100, 200, or 500 samples. This procedure is meant to provide example datasets where the model is guaranteed to hold and the true generating graph is known, but the graphs, the correlations, and the distributions are still taken from real data.

In our experiments, we have used 10 different real world data sets with 2-3 variables each from meteorological, biological, economic, and geological sources[3].

The accuracy of all the tested methods on the different datasets is shown in Figure 4. The main findings confirm the results of the synthetic data simulations: BayesLiNGAM on average outperforms the competition, although for data which is close to Gaussian the GH method can be as good or even better. Note that whereas the GL parametrization seemed better suited to the synthetic data, the MoG density model is on average superior in fitting the real distributions.

## 6 DISCUSSION

A key question is how to apply the present method to problems involving more than a handful of variables. Fortunately, as discussed in Section 3.4, the computation of the marginal likelihood of any given DAG can be broken down into smaller parts, making it feasible to score even relatively large networks, particularly so if the networks are sparsely connected. In the context of inference of Bayesian networks over discrete-valued variables, this special structure of the objective function has been widely used to find practical algorithms for finding the globally optimal DAG (see, e.g. Silander and Myllymäki (2006)) as well as computing posterior probabilities of submodels or 'features' of the full DAG (see Koivisto and Sood (2004)).

The modular structure of the objective also benefits algorithms that perform a greedy local search for the optimal DAG. Although not guaranteed to find the highest-scoring DAG for finite sample sizes, such methods can nevertheless find reasonable models with moderate computational cost.

---

[3]Climate data collected by the Deutscher Wetter Dienst (www.dwd.de), the Abalone data from the UCI repository (www.ics.uci.edu/~mlearn/MLRepository.html), the Dow Jones Index and other econometric data from (www.robjhyndman.com/TSDL/), and the Old Faithful data which is available directly in R. Full details and code for reproducing the results are provided in our code package.

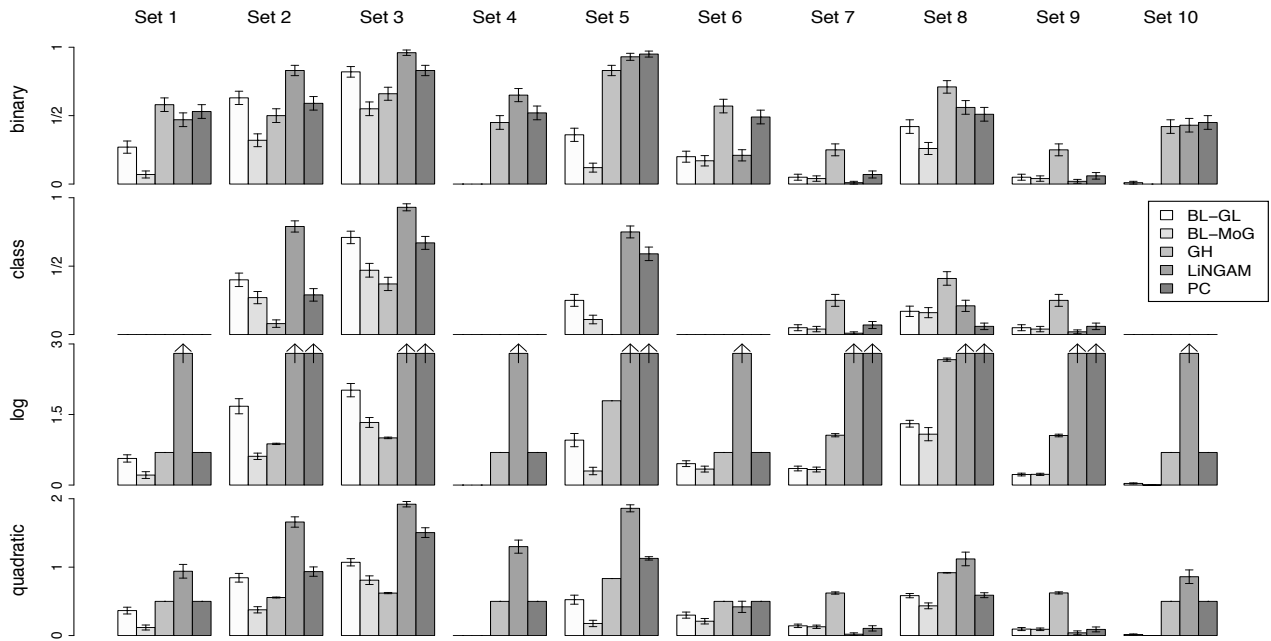

Figure 4: Results of the simulations based on real data. Shown are the average losses (rows in the array) obtained for each of the applied methods (see legend) using each of the different real datasets to generate the data (columns in the array). Lower values are better. Error bars denote the standard error on the mean (in some cases, this was too small to draw any error bars). Note that for the logarithmic measure both LiNGAM and PC often obtained infinite values because they effectively assigned zero probability to the true model in some of the experimental cases; these have been marked by an arrow rather than error bars. Empty subplots (in the 'class loss' row) resulted from easy problems for which none of the methods made any errors. See main text for discussion.

To illustrate, in Figure 5 we give an example small enough ($n = 6$ variables) that we can exhaustively compute the marginal likelihoods for all $3,781,503$ DAGs (based on $n\,2^{n-1} = 192$ different components or 'families'), find the highest-scoring graph, and moreover exactly compute the normalizing constant $p(\mathcal{D})$. Figure 5a gives the original data-generating model, in which the disturbance for variable 1 was sub-Gaussian, those for variables 4 and 5 slightly super-Gaussian, and the rest of the disturbances were Gaussian. A total of 5000 samples were simulated from the model.

We implemented a greedy local search algorithm which starts from the empty graph and iteratively evaluates all graphs obtained from the current graph by adding, removing or reversing arcs, and among these always selects the one obtaining the highest score, until no further improvement is possible. Using the MoG density model and Laplace approximation, this greedy approach to `BayesLiNGAM` converged on candidate 1 (Figure 5b). Searching all neighbors yielded candidate 2 (Figure 5c) as the only significant competing model. Summing the marginal likelihoods of all evaluated DAGs to approximate the normalizing constant $p(\mathcal{D})$ yielded posterior probabilities of 0.50 for candi-date 1 and 0.33 for candidate 2. (Exhaustive enumeration of all DAGs confirmed these two to be the main candidates and adjusted the posterior probabilities to 0.45 and 0.30, respectively.) Using the GL model and MCMC to estimate the marginal likelihood gave similar results, although the posterior probabilities were 0.71 and 0.14 respectively. On this same data, PC found the Markov equivalence class of candidate 1, while LiNGAM made several mistakes.

In summary, it should be relatively straightforward to apply techniques such as these to learn graphs over a larger number of variables, although it must be admitted that our present implementation is far from ideal in terms of computational efficiency. An interesting direction would be to combine efficient search methods such as PC or LiNGAM with the score-based approach introduced here.

Another useful extension would be to allow nonlinear relationships between the variables (with a prior which favored smooth functions), but keeping the requirement of independent additive noise. The identification of such models has already been considered in (Hoyer et al., 2009),

## 7 CONCLUSIONS

Causal discovery from non-experimental data is already a well studied topic. It is well recognized that using observational data alone it will always be impossible to strictly prove causality without strong assumptions, thus all 'causal discovery' methods rely on some principle of simplicity: If the observed data can be modeled with a 'simple' causal model in which A causes B, but require a comparatively 'complicated' model to explain B causing A, this is taken as evidence for the former case. In the method presented here, simplicity is measured both in terms of the data admitting a linear (as opposed to nonlinear) model, and in terms of the model requiring as few parameters as possible.

Although the limits of identifiability are particularly well understood for this linear and causally sufficient case, existing causal discovery methods have still been limited in their applicability with respect to the form of the distributions involved (Gaussian or non-Gaussian) and their stability with regards to the number of samples available. In this contribution, we have demonstrated a Bayesian approach that, at least for small graphs, seems to equal or outperform the existing methods in terms of accuracy. Furthermore, we provide an online code package that we hope will allow more researchers to try out these methods.

**Acknowledgements**

We wish to thank Peter Spirtes for providing the R code for the PC algorithm, available as part of our code package. The authors were supported by the Academy of Finland (project #1125272) and by University of Helsinki Research Funds (project #2105061).

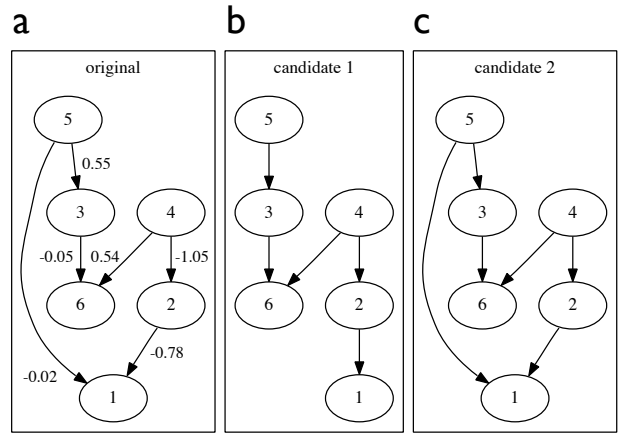

Figure 5: Synthetic data example over 6 variables. **(a)** Original generating network with weights $b_{ij}$. **(b)** The DAG maximizing the marginal likelihood (MoG parametrization, Laplace approximation). **(c)** Second-highest scoring network. See main text for details.